\documentclass{article}

\usepackage{microtype}
\usepackage{graphicx}
\usepackage{subfigure}
\usepackage{booktabs} 

\usepackage{hyperref}

\usepackage[accepted]{icml2024}

\usepackage{amsmath}
\usepackage{amssymb}
\usepackage{mathtools}
\usepackage{amsthm}

\usepackage[capitalize,noabbrev]{cleveref}

\theoremstyle{plain}

\theoremstyle{definition}

\theoremstyle{remark}

\usepackage[textsize=tiny]{todonotes}

\icmltitlerunning{EPD: Long-term Memory Extraction, Context-awared Planning and Multi-iteration Decision @ EgoPlan Challenge ICML 2024}

\begin{document}

\twocolumn[
\icmltitle{EPD: Long-term Memory Extraction, Context-awared Planning and Multi-iteration Decision @ EgoPlan Challenge ICML 2024}



\icmlsetsymbol{equal}{*}
\icmlsetsymbol{correspond}{\dag}

\begin{icmlauthorlist}
\icmlauthor{Letian Shi}{yyy}
\icmlauthor{Qi Lv}{yyy,sch1,sch2}
\icmlauthor{Xiang Deng}{correspond,yyy}
\icmlauthor{Liqiang Nie}{correspond,yyy}
\end{icmlauthorlist}

\icmlaffiliation{yyy}{Harbin Institute of Technology (Shenzhen)}
\icmlaffiliation{sch1}{School of Engineering, Great Bay University}
\icmlaffiliation{sch2}{School of Computing and Information Technology, Great Bay University}

\icmlcorrespondingauthor{Xiang Deng}{dengxiang@hit.edu.cn}
\icmlcorrespondingauthor{Liqiang Nie}{nieliqiang@gmail.com} 

\icmlkeywords{Machine Learning, ICML}

\vskip 0.3in
]



\printAffiliationsAndNotice{\CorrespondingAuthor}

\begin{abstract}

In this technical report, we present our solution for the EgoPlan Challenge in ICML 2024. To address the real-world egocentric task planning problem, we introduce a novel planning framework which comprises three stages: long-term memory \textbf{E}xtraction, context-awared \textbf{P}lanning, and multi-iteration \textbf{D}ecision, named \textbf{EPD}.
Given the task goal, task progress, and current observation, the extraction model first extracts task-relevant memory information from the progress video, transforming the complex long video into summarized memory information. The planning model then combines the context of the memory information with fine-grained visual information from the current observation to predict the next action. Finally, through multi-iteration decision-making, the decision model comprehensively understands the task situation and current state to make the most realistic planning decision. On the EgoPlan-Test set, EPD achieves a planning accuracy of 53.85\% over 1,584 egocentric task planning questions. We have made all codes available at \href{https://github.com/Kkskkkskr/EPD}{https://github.com/Kkskkkskr/EPD}.
\end{abstract}

\section{Introduction}
\label{Introduction}

Egocentric task planning is a crucial aspect of human activity, requiring the perception and understanding of the world from the egocentric perspective to make informed decisions in complex environments and solve real-world problems. This kind of task has broad research prospects in applications such as robotics. To accurately and comprehensively evaluate the egocentric task planning capabilities of Multimodal Large Language Models (MLLM), the EgoPlan-Bench~\cite{chen2023egoplan} has been proposed. It is built based on egocentric real-world video datasets from Epic-Kitchens~\cite{damen2022rescaling} and Ego4D~\cite{grauman2022ego4d}. To correctly answer the task planning questions in EgoPlan, models must comprehend real-world task objectives, track the progress of long-term tasks, perceive the environment from an egocentric perspective, and infer the next steps based on their own knowledge. This places new demands on the models' abilities to process long videos and understand complex context.

Currently, most existing work is based on a single visual modality, adapting video models through fine-tuning to bridge the domain gap between exocentric videos and egocentric videos, transforming them into planners capable of real-world egocentric task planning. This approach leverages the strong long-video processing capabilities of video models and captures fine-grained visual information in videos through fine-tuning for task planning. However, the training process of video models is highly time-consuming and resource-intensive, which brings limitations. On the other side, a common approach is designing templates to prompt large language models (LLMs)~\cite{wang2024videoagent, zhang2023simple}, leveraging their powerful understanding abilities. Although this method often faces challenges due to the performance limitations of LLMs, making it difficult to perfectly adapt to the specific characteristics of downstream tasks, it requires significantly less time and resources and can fully utilize the extensive knowledge embedded in LLMs. Therefore, in this challenge, we decide to design our solution by designing templates to prompt the MLLM.

Unlike long video summarization and visual question answering, egocentric task planning requires the model to comprehensively understand the task goals, real-time progress, and current observations, as well as the relationships between them. For example, when the task goal is ``Add garlic to the food and stir" and the chopping board is currently held in hand, the model needs to select ``add garlic to food" to advance the task, rather than choosing to repeat ``pick up chopping board" simply because a similar action appears in the progress video. 
In egocentric task planning, despite the progress video comprising a large proportion of the visual input, it holds equal importance to the current observation represented as a single image.
Therefore, the model must fully understand the meaning and connections between task objectives, progress, and current state to effectively plan the next action. This presents significant challenges for current MLLMs.

To address these issues, we propose a novel long-term memory \textbf{E}xtraction, context-awared \textbf{P}lanning, and multi-iteration \textbf{D}ecision framework (\textbf{EPD}) , which tackles the aforementioned problems through three stages: (1) long-term memory extraction: To extract core action information from lengthy video frames, we design a long-term memory extraction mechanism, leveraging MLLMs to understand, extract, and memorize key content from continuous clip segments; (2) context-awared planning: To enable the planning model to comprehensively understand the context and make informed plans, we use memory information along with current observations as multimodal inputs and design corresponding prompts to assist MLLMs in planning; (3) multi-iteration decision: Given the difficulty for MLLMs to capture all details in a single attempt, we employ a multi-iteration decision-making process, allowing the decision model to compare different planning processes and determine the most reasonable plan as the answer. Notably, the EPD framework we propose is training-free, so our method does not require additional training time and computational resources. Through this training-free approach, we achieves a planning accuracy of 53.85\% on the EgoPlan-Test set, which is approximately a 10\% improvement over the baselines.

\begin{figure*}[h]
\vskip 0.2in
\begin{center}
\centerline{\includegraphics[width=\textwidth]{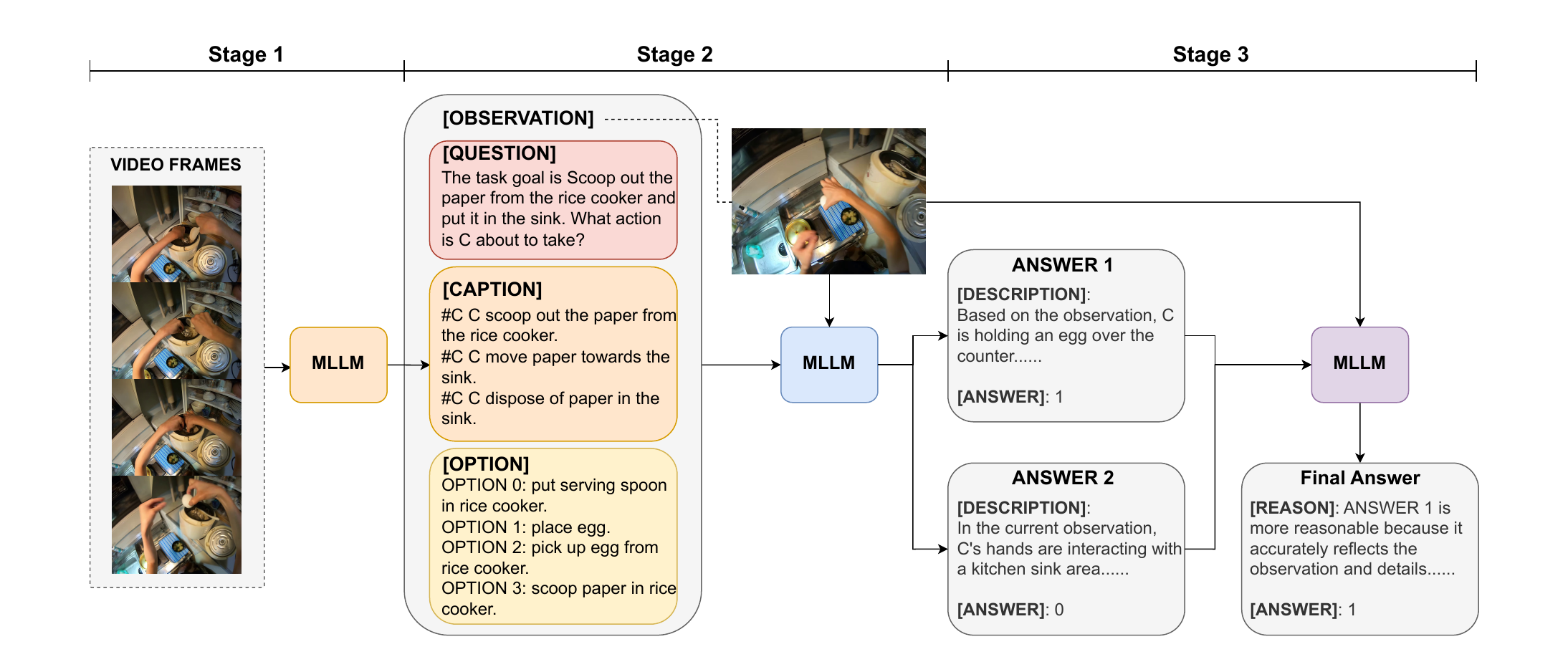}}
\caption{Framework of our solution EPD.}
\label{fig:framework}
\end{center}
\vskip -0.2in
\end{figure*}

\section{Methodology}
As shown in the \textit{Figure} \hyperref[fig:framework]{\ref*{fig:framework}}, given a process video, memory extraction model first extracts the video content, excluding the current observation, into memory text information. The memory text information, along with the current observation, is then input to the planning model to plan the next action. Finally, multi-iteration decision is employed to allow the decision model to select the most reasonable answer from different answers.

\subsection{Long-term Memory Extraction} 
The videos provided by EgoPlan consist of multiple action segments within specified time frames. To obtain the memory text information for each action, we extract four frames from each action video segment, including the first and last frames and two evenly selected frames within the segment, and input these frames into a memory extraction model. This model needs to accurately extract the action content of the segment in a single sentence. For model selection, we initially try the existing video processing model LaViLa~\cite{zhao2023learning} and VideoRecap~\cite{islam2024video}, but they have fail to accurately recognize objects, resulting in missing object descriptions. Ultimately, we choose the GPT-4o model for memory extraction.

\subsection{Context-awared Planning} 
Once the memory text information is obtained, it is input into the planning model as task process descriptions, transforming the complex fine-grained video understanding into textual planning. We use both the memory text information and the current visual observation as multimodal input, prompting the task planning model through in-context learning to output the next-step planning. In this process, the task planning model includes both GPT-4o and Claude 3.5. We use different models to obtain different results, which serve as input for subsequent multi-iteration decision.

\subsection{Multi-iteration Decision} 
Through multi-iteration decision, we can leverage the comprehensive reasoning ability of GPT-4o and the fine-grained textual reasoning ability of Claude 3.5, allowing the two models to complement each other's performance. During the decision process, we provide the different planning results from GPT-4o and Claude 3.5 to the decision model, which compares the different planning processes and decides the most reasonable answer. Since GPT-4o achieves the best single-model results, we choose GPT-4o as the model used in the decision process. We also attempt multi-iteration decision by selecting the most frequent option among multiple answers, and the results are close to the optimal decision results obtained through multi-iteration decision with GPT-4o.

\section{Experiment}
\subsection{Performance Comparison}
\begin{table*}[t]
\caption{Performance Comparison on EgoPlan-Val OD(out-of-domain) and EgoPlan-Test}
\label{tab:comparison}
\vskip 0.15in
\begin{center}
\begin{small}
\begin{sc}
\begin{tabular}{lcccr}
\toprule
Model & EgoPlan-Val OD & EgoPlan-Test & training-free \\
\midrule
Video-LLaMA           & 30.44 & 30.30          & $\surd$ \\
GPT-4V                & 36.90 & 37.25          & $\surd$ \\
Enhanced Video-LLaMA  & 44.42 &                & $\times$\\
EPD                   &       & \textbf{53.85} & $\surd$ \\

\bottomrule
\end{tabular}
\end{sc}
\end{small}
\end{center}
\vskip -0.1in
\end{table*}

According to the current leaderboard, the existing results are shown in \textit{Table} \hyperref[tab:comparison]{\ref*{tab:comparison}}. On the EgoPlan-Val out-of-domain set, the best training-free baseline is GPT-4V, with an accuracy of 36.90\%. The best trained baseline, enhanced Video-LLaMA, is obtained by fine-tuning Video-LLaMA~\cite{zhang2023video} on the EgoPlan dataset and achieves an accuracy of 44.42\%. Our method EPD achieves an accuracy of 53.85\% on the EgoPlan-Test set, significantly surpassing all previous baselines. This demonstrates the effectiveness of our approach.

\subsection{Solution Evolution}

\begin{table*}[t]
\caption{Solution Evolution}
\label{tab:evolution}
\vskip 0.15in
\begin{center}
\begin{small}
\begin{sc}
\begin{tabular}{lcccr}
\toprule
Order & Accuracy & Modification \\
\midrule
1    & 41.28            & Use LaViLa to extract memory \\
2    & 45.58            & Use GPT-4o to extract memory \\
3    & 48.04            & Adopt four-shot in-context learning \\
4    & 51.83            & Use high-resolution observation and output description \\
5    & 52.71            & Multi-iteration Decision with five GPT-4o results\\
6    & \textbf{53.85}   & Multi-iteration Decision with GPT-4o and Claude-3.5 results \\

\bottomrule
\end{tabular}
\end{sc}
\end{small}
\end{center}
\vskip -0.1in
\end{table*}

The iterative optimization process of our solution is illustrated in \textit{Table} \hyperref[tab:evolution]{\ref*{tab:evolution}}. Initially, we compare the overall framework performance using LaViLa and GPT-4o as memory extraction models. The results demonstrate a 4.3\% performance improvement with GPT-4o over LaViLa, as GPT-4o produces memory text information with more accurate object descriptions.
Subsequently, in our prompt research, we identify that insufficient examples lead to inadequate understanding of the planning and reasoning process. To address this, we adopt four-shot in-context learning, carefully selecting four examples from the validation dataset. We extract memory text information and manually complete the reasoning steps, achieving a 2.46\% performance improvement.
Throughout our experiments, the importance of the current observation becomes increasingly evident. By changing the input of the current observation to a high-resolution original image and prompting GPT-4o to first describe its visual content in text, we enhance the model's understanding of the current observation, resulting in a 3.79\% performance improvement.
During the multi-iteration decision phase, we initially try a simple option counting method, selecting the most frequently occurring option from five results, which improves performance by 0.88\% compared to a single result. In further exploration of multi-iteration decision, we use GPT-4o to decide whether GPT-4o's planning or Claude 3.5's planning is more convincing, selecting the more reasonable answer. This method improves performance by 1.14\% over the previous decision approach, achieving the optimal accuracy result of our solution.

\subsection{Ablation Study}
\begin{table*}[t]
\caption{Ablation Study}
\label{tab:ablation}
\vskip 0.15in
\begin{center}
\begin{small}
\begin{sc}
\begin{tabular}{lcccr}
\toprule
 Planning Model & Input Video with Memory Information &  Accuracy  \\
\midrule
 Claude 3.5  & $\times$ & 47.22 \\
 GPT-4o      & $\surd$  & 49.68 \\
 GPT-4o      & $\times$ & \textbf{51.83} \\
\bottomrule
\end{tabular}
\end{sc}
\end{small}
\end{center}
\vskip -0.1in
\end{table*}
We examine the impact of different planning models, and the inclusion of video input on the results, as shown in \textit{Table} \hyperref[tab:ablation]{\ref*{tab:ablation}}. It is evident that GPT-4o exhibits stronger multimodal planning capabilities on EgoPlan compared to Claude 3.5. However, we also find that although training-free models require less time and resources, they cannot attain improved fine-grained egocentric visual understanding capabilities by training on the EgoPlan dataset. We attribute this to the inability of simple in-context learning methods to achieve complex visual understanding.

\begin{figure}[ht]
\vskip 0.2in
\begin{center}
\centerline{\includegraphics[width=\columnwidth]{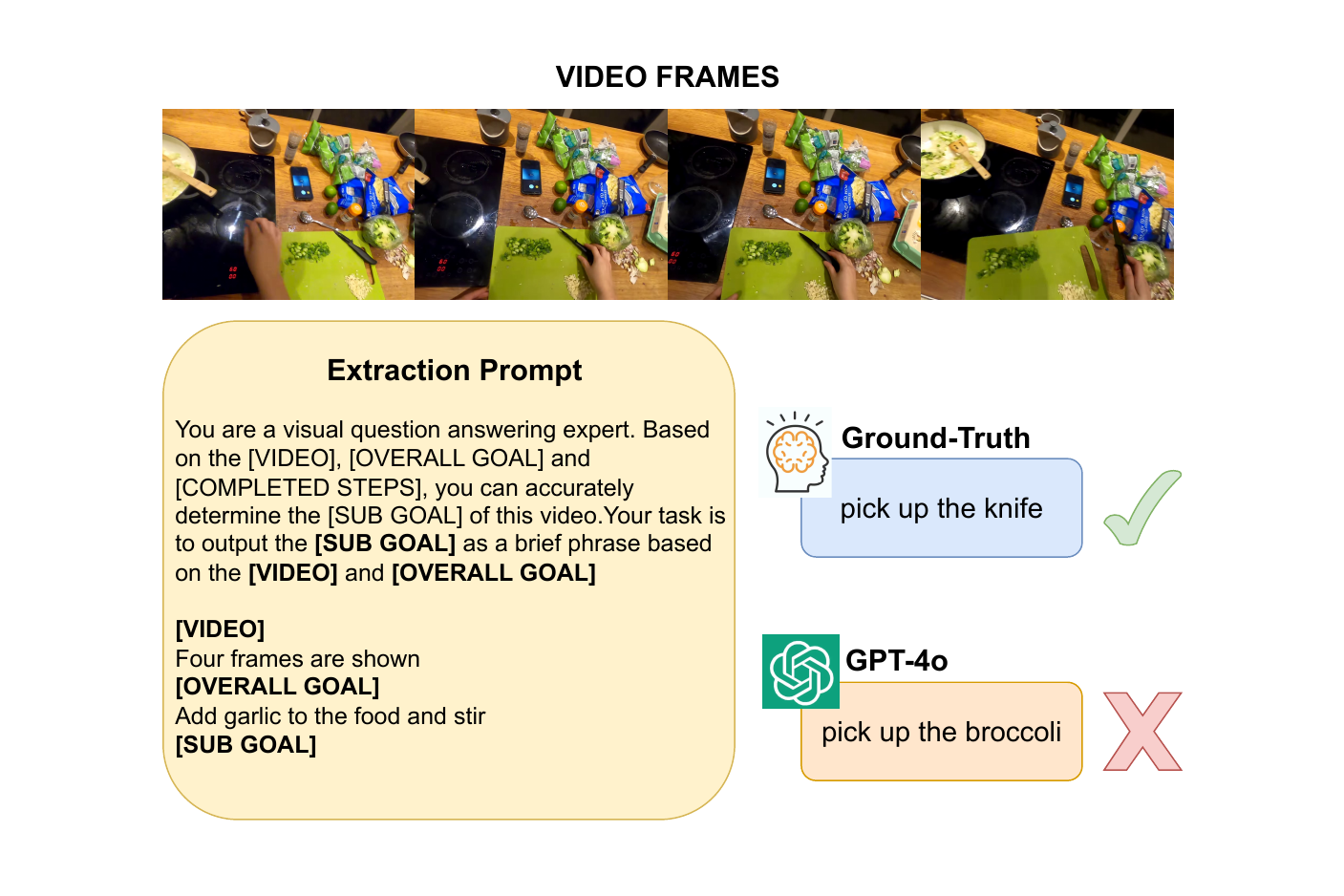}}
\caption{Wrong Extraction with GPT-4o}
\label{fig:casestudy}
\end{center}
\vskip -0.2in
\end{figure}

\subsection{Case Analysis}
Although GPT-4o exhibits strong performance as a multimodal large language model, it still lacks fine-grained visual understanding. As shown in the \textit{Figure} \hyperref[fig:casestudy]{\ref*{fig:casestudy}}, despite the action being ``pick up knife", GPT-4o incorrectly interprets it as ``pick up broccoli". This aligns with the key issue discussed in EgoPlan-Bench: although our solution EPD achieves low time and resource consumption in the EgoPlan Challenge, fine-grained visual understanding on real-world egocentric datasets can only be attained through further training.

From this perspective, the problem posed by EgoPlan is highly significant in practical terms. Additionally, trainable models have substantial potential in addressing the EgoPlan task, with considerable room for improvement.

\section{Conclusion}
This paper proposes a real-world egocentric task planning framework, EPD, which achieves a planning accuracy of 53.85\% in the EgoPlan Challenge in ICML 2024, demonstrating the effectiveness of this approach. Additionally, we identify the limitations of training-free models in capturing fine-grained visual semantics, providing new insights for improving trainable models on EgoPlan in future research.

\nocite{chen2023egoplan}
\nocite{damen2022rescaling}
\nocite{grauman2022ego4d}
\nocite{wang2024videoagent}
\nocite{zhang2023simple}
\nocite{zhao2023learning}
\nocite{islam2024video}
\nocite{zhang2023video}

\bibliography{example_paper}
\bibliographystyle{icml2024}

\newpage
\appendix
\onecolumn



\end{document}